\documentclass[11pt]{article}

\usepackage[preprint]{acl}

\usepackage{times}
\usepackage{latexsym}
\usepackage{amsmath}
\usepackage{amssymb}

\usepackage[T1]{fontenc}
\usepackage[utf8]{inputenc}

\usepackage{microtype}
\usepackage{xurl}

\usepackage{inconsolata}

\usepackage{graphicx}

\usepackage{booktabs}  % For professional tables
\usepackage{enumitem}  % For compact lists
\usepackage{array}

\usepackage{xcolor}
\definecolor{entitycolor}{RGB}{0,123,255}

\title{Just Pass Twice: Efficient Token Classification with LLMs for Zero-Shot NER}

\author{
  \textbf{Ahmed Ewais}\textsuperscript{\dag} \quad
  \textbf{Ahmed Hashish}\textsuperscript{\dag} \quad
  \textbf{Amr Ali} \\[0.5em]
  WitnessAI \\[0.3em]
  \texttt{\{ahmed, ahmed.hashish, amr\}@witness.ai}
}

\begin{document}
\maketitle
\begingroup\renewcommand{\thefootnote}{}
\footnotetext{\textsuperscript{\dag}Equal contribution.}
\footnotetext{Project page: \url{https://witness-ai-jpt-ner.hf.space/}}
\endgroup
\begin{abstract}

Large language models encode extensive world knowledge valuable for zero-shot named entity recognition. However, their causal attention mechanism, where tokens attend only to preceding context, prevents effective token classification when disambiguation requires future context. Existing approaches use LLMs generatively, prompting them to list entities or produce structured outputs, but suffer from slow autoregressive decoding, hallucinated entities, and formatting errors.

We propose \textbf{Just Pass Twice (JPT)}, a simple yet effective method that enables causal LLMs to perform discriminative token classification with full bidirectional context. Our key insight is that concatenating the input to itself lets each token in the second pass attend to the complete sentence, requiring no architectural modifications. We combine these representations with definition-guided entity embeddings for flexible zero-shot generalization.
Our approach achieves state-of-the-art results on zero-shot NER benchmarks, surpassing the previous best method by \textbf{+7.9 F1} on average across CrossNER and MIT benchmarks, being over 20$\times$ faster than comparable generative methods.
Code and pretrained model weights will be available through our \href{https://witness-ai-jpt-ner.hf.space/}{project page}.
\end{abstract}

\begin{figure}[t]
    \centering
    \includegraphics[width=\columnwidth]{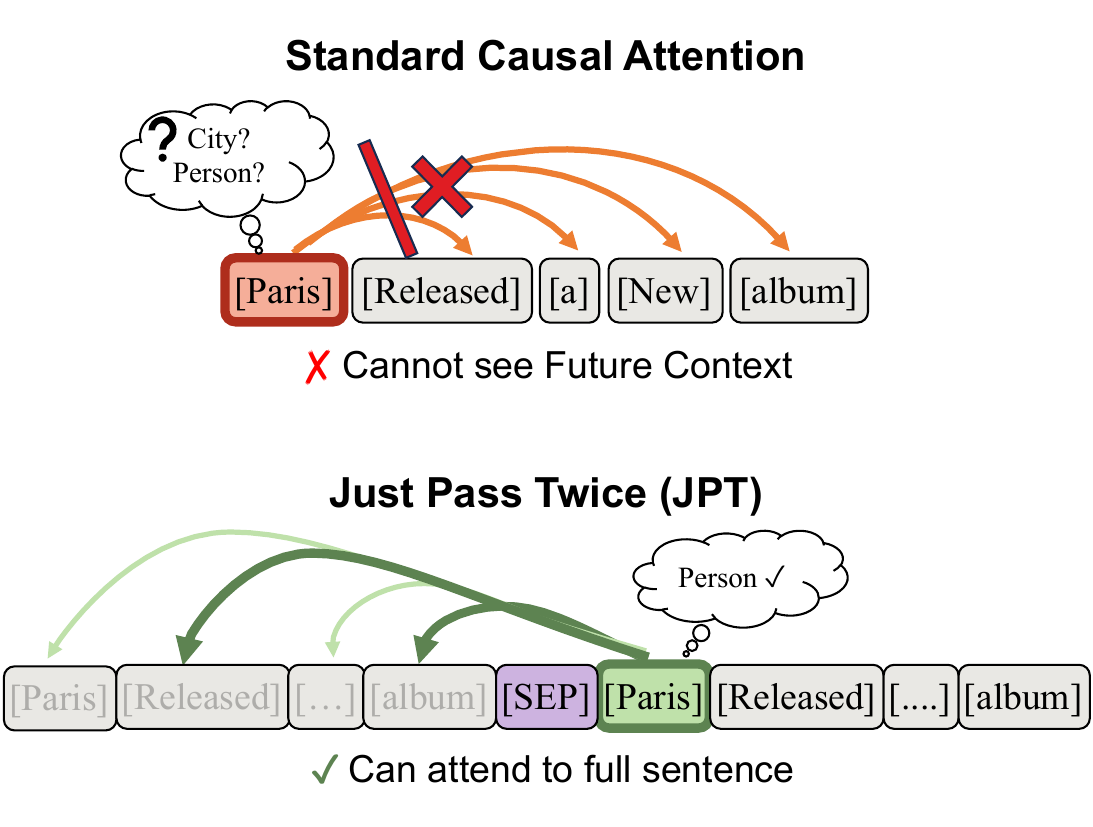}
    \setlength{\abovecaptionskip}{4pt}

    \caption{\textbf{Just Pass Twice (JPT) enables bidirectional token classification in causal LLMs.}
    \textbf{(Top)} Standard causal masking restricts the target token ``Paris'' from attending to future context like ``album'' (red barrier), leading to ambiguity.
    \textbf{(Bottom)} JPT duplicates the input sequence. In the second pass, the target token (green box) attends backwards to the \textit{entire} original sequence (green arrows), resolving the entity type without architectural modifications.}
    
    \label{fig:teaser}
    \vspace{-1em} % Tightens the whitespace between the caption and the main body text
\end{figure}

\section{Introduction}

Named Entity Recognition (NER) is a foundational natural language processing task, underpinning numerous downstream applications such as information extraction, privacy-preserving text processing, knowledge graph construction, entity linking, question answering, and document understanding pipelines \citep{keraghel2024recentadvancesnamedentity}. Early approaches framed NER as a token-level sequence labeling problem, most commonly using the BIO tagging scheme \citep{chiu-nichols-2016-named,akbik-etal-2018-contextual,qin-etal-2019-stack,devlin-etal-2019-bert}, where every token is assigned a label indicating whether it begins, is inside, or is outside an entity mention.

The dominant approach to NER has been discriminative token classification using bidirectional encoders such as BERT \citep{devlin-etal-2019-bert}, RoBERTa \citep{liu2019robertarobustlyoptimizedbert}, and DeBERTa \citep{he2021debertadecodingenhancedbertdisentangled}. This paradigm underlies a wide range of NER systems, including BioNER based on BERT \citep{cocchieri-etal-2025-openbioner}, RoBERTa-based approaches such as NuNER \citep{bogdanov-etal-2024-nuner}, and DeBERTa-based systems such as GLiNER \citep{zaratiana-etal-2024-gliner}. These models naturally support token-level labeling: their bidirectional attention allows each token to attend to both preceding and subsequent context, enabling effective disambiguation. However, encoder-based methods face inherent limitations: they are typically small ($<$1B parameters), have limited context windows, and encode less world knowledge than larger models, which can limit generalization to novel entity types and domains.

Large language models (LLMs) offer a compelling alternative. With billions of parameters and training on vast corpora, LLMs encode rich world knowledge and demonstrate remarkable reasoning capabilities across diverse NLP tasks. These properties make them seemingly ideal for zero-shot NER.

However, existing approaches to LLM-based NER diverge from the sequence labeling paradigm, instead employing generative formulations. Recent work has fine-tuned open-source LLMs on diverse NER datasets to enhance domain adaptability: InstructUIE \citep{wang2023instructuiemultitaskinstructiontuning} trains on a wide range of information extraction datasets using instruction tuning; UniversalNER \citep{zhou2024universalner} distills from ChatGPT and queries entity types one at a time for improved recall; and GoLLIE \citep{sainz2024gollie} uses code-style annotation guidelines to improve zero-shot generalization. Empirical evaluations show that even ``vanilla'' prompting of large models like ChatGPT yields suboptimal results compared to smaller supervised baselines \citep{wei2024chatiezeroshotinformationextraction,li2023evaluatingchatgptsinformationextraction}.

Despite these advances, all generative NER approaches suffer from fundamental drawbacks:
(1) \textbf{Speed}: autoregressive token generation is inherently slow compared to single-pass classification;
(2) \textbf{Cost}: output tokens cost 3--4$\times$ more than input tokens in commercial APIs because decoding is sequential and memory-bound, requiring a forward pass per generated token; (3) \textbf{Hallucinations and Format Errors}: Generative models are prone to hallucinating entities not present in the input and can produce outputs that fail to parse into valid structured data, requiring error handling or re-prompting \citep{li2023evaluatingchatgptsinformationextraction,wang-etal-2025-gpt}.

A natural question arises: why not apply the successful token classification paradigm from encoders directly to LLMs? The fundamental obstacle is architectural. Modern LLMs employ causal (unidirectional) attention, where each token can only attend to itself and preceding tokens. While essential for efficient autoregressive generation, this creates a critical limitation for token classification: when classifying a token, the model lacks access to subsequent context that may be essential for disambiguation.

Consider the sentence \emph{``Paris released a new album''} (Figure~\ref{fig:teaser}). To correctly classify \emph{Paris} as a person rather than a location, a model must see the full sentence; the noun phrase \emph{``a new album''} reveals this refers to the musician, not the city. Under causal attention, when the model processes \emph{Paris}, it has not yet observed \emph{released}, \emph{new}, or \emph{album}. The causal mask prevents the classifier from accessing this disambiguating future context, which is precisely why standard token classification fails with decoder-only LLMs.

We propose Just Pass Twice (JPT), a method that bridges this gap through two key innovations. \textbf{First, we enable bidirectional token classification in causal LLMs without architectural changes.} Our insight is simple: concatenating the input sequence to itself allows each token in the second pass to attend to the complete sentence from the first pass. While recent works duplicate inputs and pool the sequence into a single global vector to improve text embeddings for retrieval tasks \citep{springer-etal-2025-repetition}, JPT leverages duplication specifically to bypass the causal attention bottleneck, extracting unpooled, bidirectional hidden states for individual tokens to perform fine-grained NER. Despite doubling the input length, the operation occurs entirely in the highly parallel prefill phase. Unlike autoregressive decoding, which is sequential and memory-bound, this makes JPT dramatically faster than generative alternatives.

\textbf{Second, we introduce definition-guided entity typing for flexible zero-shot generalization.} Rather than encoding entity types by name alone (e.g., ``PERSON''), we encode rich natural language definitions that precisely specify what each type encompasses. We inject definitions through two complementary channels: (1) as embeddings that the classifier matches against token representations, and (2) directly in the LLM's input prompt, where they guide token encoding via attention. Inspired by recent work showing that definitions outperform simple label names for zero-shot NER \citep{cocchieri-etal-2025-openbioner}, this approach decouples the model from any fixed label vocabulary while providing fine-grained control: users can specify boundary cases directly in natural language, such as whether \texttt{PRICE} captures only explicit amounts (``\$50'') or also qualitative descriptors (``budget-friendly''). Since definitions remain fixed at inference time, entity embeddings are computed offline and cached, adding no runtime overhead.

We implement JPT by adding lightweight LoRA adapters \citep{hu2021loralowrankadaptationlarge}, projection layers, and a bilinear classifier to frozen Qwen3 backbones (4B and 8B parameters), trained on an in-house Wikipedia-derived NER dataset with no overlap with evaluation benchmarks. On zero-shot evaluation across CrossNER and MIT benchmarks, JPT surpasses state-of-the-art methods by \textbf{+7.9 F1} on average, while being over 20$\times$ faster than comparable generative methods. We further demonstrate JPT's strong generalization on an extended benchmark of 20 diverse NER datasets spanning biomedical, social media, and multilingual domains.

Our main contributions are:
\begin{itemize}
    \item A simple, effective method for enabling bidirectional context in causal LLMs through input duplication, requiring no architectural modifications to the base LLM architecture and leveraging efficient parallel prefill computation.
    \item Definition-guided entity typing that enables flexible zero-shot generalization through natural language type specifications, offering fine-grained control over what constitutes each entity type.
    \item State-of-the-art results on CrossNER and MIT benchmarks (+7.9 F1 over the previous best), with consistent improvements across 19 of 20 extended benchmarks and over 20$\times$ speedup versus generative methods.
\end{itemize}

\section{Related Work}
\label{sec:related}

The landscape of zero-shot Named Entity Recognition is characterized by a fundamental tension: efficient discriminative models with limited capacity and reasoning capabilities versus powerful generative LLMs with high latency and reliability issues.
\subsection{Generative NER with LLMs}
The dominant paradigm for LLM-based NER formulates the task as sequence generation. \textbf{UniversalNER} \citep{zhou2024universalner} demonstrated that distilling ChatGPT into smaller generative models (e.g., LLaMA-7B) via targeted instruction tuning, querying one entity type at a time, can achieve impressive open-vocabulary performance. \textbf{InstructUIE} \citep{wang2023instructuiemultitaskinstructiontuning} established a unified framework showing that multi-task instruction tuning captures inter-task dependencies. To handle complex schema definitions, \textbf{GoLLIE} \citep{sainz2024gollie} fine-tunes models to follow annotation guidelines formatted as code, improving zero-shot generalization to unseen schemas. Building on these instruction-tuning paradigms, \textbf{GNER} \citep{ding-etal-2024-rethinking} identifies that previous methods are overly ``entity-centric''; they propose incorporating negative instances (non-entities) into training to explicitly refine boundary detection and context awareness. \textbf{SaM} \citep{ding-etal-2025-selecting} dynamically selects and merges domain-specific LoRA adapters at inference time, achieving strong zero-shot performance but requiring a library of pre-trained expert weights.

However, generative approaches face inherent limitations. \textbf{GPT-NER} \citep{wang-etal-2025-gpt} identified the ``hallucination issue,'' where LLMs over-confidently generate entities not present in the input, necessitating secondary verification steps. More fundamentally, autoregressive decoding introduces latency that scales with output length rather than input length.

Some recent approaches seek to improve accuracy by prompting LLMs for explicit explanations or justifications. For example, \textbf{PromptNER} \citep{ashok2023promptnerpromptingnamedentity} asks models to produce explanations supporting entity compatibility. While such methods can bolster interpretability and performance, they further increase generation costs and latency by requiring models to output rationales in addition to entity predictions. JPT sidesteps this trade-off entirely: we leverage the underlying reasoning capacity of LLM backbones but bypass autoregressive generation through discriminative projection.

\subsection{Discriminative Encoder-Based Approaches}
Parallel to generative methods, discriminative architectures treat NER as a semantic matching problem. 
\textbf{GLiNER} \citep{zaratiana-etal-2024-gliner} uses a bidirectional transformer (DeBERTa) to encode concatenated entity type prompts and text, enabling zero-shot detection via span-type matching in a shared latent space.
\textbf{OpenBioNER} \citep{cocchieri-etal-2025-openbioner} extends this paradigm using a cross-encoder architecture tailored to the biomedical domain, demonstrating that encoding entity \emph{definitions}, rather than simple label names, significantly boosts zero-shot performance on rare concepts. \textbf{NuNER} \citep{bogdanov-etal-2024-nuner} pushes the encoder paradigm further by employing a bi-encoder architecture (separating text and concept encoding) pretrained on massive synthetic datasets annotated by LLMs.

While these encoder-based methods are efficient, they are inherently limited by the capacity of their backbone models (typically BERT/DeBERTa at $<$1B parameters), which may encode less world knowledge than larger decoder models. JPT bridges this gap: we adopt the definition-augmented matching strategy of OpenBioNER but apply it to much larger decoder-only LLMs, while also injecting definitions directly into the prompt for dual-channel guidance.

\paragraph{Alternative Approaches to Bidirectional Context.}
Several techniques exist for obtaining bidirectional context in language models. Prefix-LM attention \citep{raffel2020exploring,tay2023ul2unifyinglanguagelearning} allows bidirectional attention over a designated prefix, but requires this pattern during pretraining. Fill-in-the-middle training \citep{bavarian2022efficienttraininglanguagemodels} enables conditioning on future context, but only for specially-trained models. Simply removing the causal mask at inference fails because attention patterns become out-of-distribution. In contrast, JPT's input duplication works with any off-the-shelf causal LLM: the model processes a standard causal sequence, but by repeating the input, tokens in the second pass attend to the complete sentence, requiring no architectural changes and no special pretraining.

\subsection{Input duplication and multi-prompt inference beyond NER}
\label{sec:duplication-related}

Recent work suggests that duplicating an input, or querying the same example through multiple prompts, can improve model behavior even outside token-level sequence labeling. For example, while \citet{springer-etal-2025-repetition} pool the second occurrence of a repeated sequence into a single vector to generate global, sentence-level embeddings for retrieval tasks, JPT fundamentally differs by extracting unpooled, bidirectional hidden states for individual tokens to enable token classification for NER.

A related line of work reuses the same input across multiple prompt formulations and aggregates the resulting outputs or predictive distributions. \citet{tonolini-etal-2024-bayesian} use semantically equivalent prompts to improve uncertainty estimation and calibration for black-box LLMs, while \citet{heineman-etal-2024-improving} and \citet{guo-etal-2025-ped} show that multi-prompt decoding can improve generation quality by increasing candidate diversity or ensembling token probabilities across prompt variants. Beyond generation, \citet{cheng-etal-2025-multi} apply multi-prompt decoding to few-shot language understanding, and \citet{lin-etal-2025-gems} use multi-perspective prompts with output aggregation for event argument extraction.

Unlike these works, which primarily target embeddings, calibration, decoding, or structured extraction through multiple prompt-conditioned views of the same input, JPT applies intra-sequence input duplication directly to discriminative token-level NER in decoder-only LLMs within a single parallel forward pass.

\subsection{Positioning Our Approach}

JPT occupies a unique position in this landscape. Unlike generative models, we operate as a discriminative token classifier, eliminating autoregressive latency and generation-related hallucinations. Unlike encoder-only models, we leverage the massive parameter space and world knowledge of 7B+ decoder LLMs. And unlike multi-stage pipelines, we require only a single forward pass. Our approach demonstrates that the causal attention constraint can be overcome through a simple input transformation rather than architectural modifications or complex inference procedures.

\section{Method}
\label{sec:method}

\subsection{Overview}

\begin{figure*}[t]
    \centering
    \includegraphics[width=\textwidth]{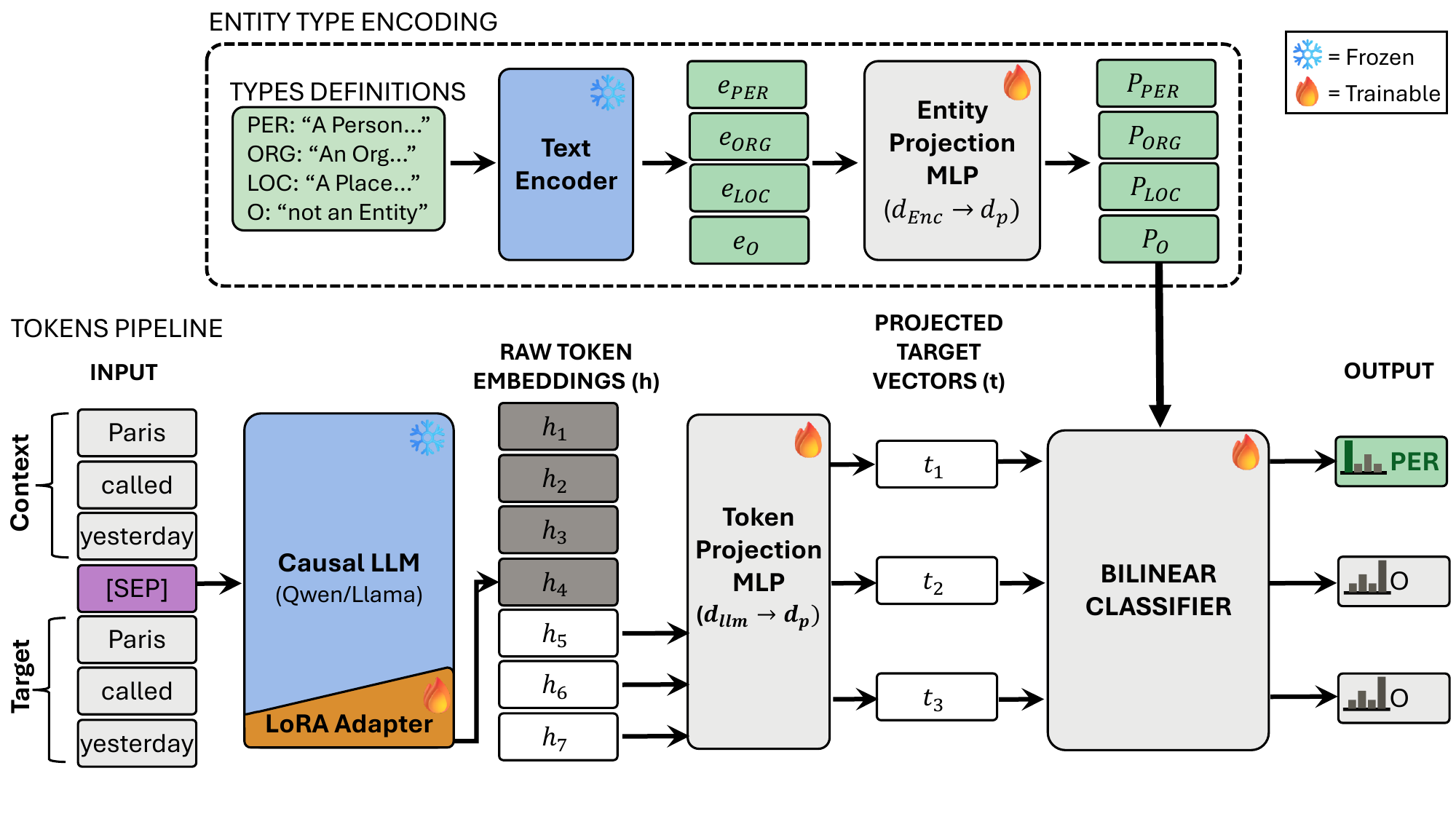}
    \caption{\textbf{Architecture of the proposed model.} 
    \textbf{(Top)} Entity definitions are encoded via a text encoder (dimension $d_{\text{enc}}$) and projected into the shared space $\mathbb{R}^{d_p}$ using the Entity Projection MLP, yielding entity embeddings $\mathbf{p}_{\text{per}}, \mathbf{p}_{\text{loc}}, \ldots$
    \textbf{(Bottom)} The Causal LLM processes the duplicated input sequence. 
    Hidden states from the second pass ($\mathbf{h}_5, \ldots, \mathbf{h}_7$) are projected to token embeddings $\mathbf{t}_1, \ldots, \mathbf{t}_n \in \mathbb{R}^{d_p}$ and scored against entity embeddings via a bilinear classifier.}
    \label{fig:architecture}
\end{figure*}

Figure~\ref{fig:architecture} illustrates the JPT architecture. Given an input text and a set of entity types with definitions, JPT performs the following steps: (1) entity type definitions are encoded using a text embedding model (this can be done offline and cached), (2) the text is duplicated and processed by a causal LLM to obtain bidirectional token representations, (3) both representations are projected to a shared space $\mathbb{R}^{d_p}$, and (4) a bilinear classifier computes matching scores between tokens and entity types.

\subsection{Bidirectional Context via Input Duplication}
\label{sec:two-pass}

The core insight of JPT is that causal attention's unidirectional constraint can be overcome through input duplication. In a causal LLM, each token $x_i$ can only attend to preceding tokens $x_1, \ldots, x_{i-1}$. This prevents effective token classification, as disambiguating context often appears \emph{after} the token of interest.

Given an input sequence $\mathbf{x} = (x_1, x_2, \ldots, x_n)$, we construct the duplicated input:
\begin{equation}
\mathbf{x}' = (\underbrace{x_1, \ldots, x_n}_{\text{first pass}}, \texttt{[SEP]}, \underbrace{x_1, \ldots, x_n}_{\text{second pass}})
\label{eq:duplication}
\end{equation}

When the LLM processes $\mathbf{x}'$, each token $x_i$ in the second pass can attend to: (1) all tokens from the first pass $x_1, \ldots, x_n$, providing complete ``future'' context, and (2) preceding tokens in the second pass $x_1, \ldots, x_{i-1}$, providing standard ``past'' context. The combination provides effective bidirectional attention without modifying the causal attention mechanism.

\paragraph{Attention Coverage.}
Crucially, for a token at position $k$ in the second pass (position $n + 1 + k$ in $\mathbf{x}'$), causal attention permits attending to all positions $j \leq n+1+k$, which includes all $n$ tokens of the first pass. This means \emph{every} token in the second pass has access to the complete input sequence, exactly the bidirectional context required for accurate token classification. We visualize these attention patterns in Section~\ref{sec:analysis}.

\paragraph{Token Representation Extraction.}
We extract hidden states only from the second occurrence of tokens, as these contain the bidirectional context. Let $\mathbf{h}_i \in \mathbb{R}^{d_{\text{llm}}}$ denote the final-layer hidden state of token $x_i$ from the second pass. We apply a learned projection:
\begin{equation}
\mathbf{t}_i = \text{MLP}_{\text{token}}(\mathbf{h}_i) \in \mathbb{R}^{d_p}
\end{equation}
where the MLP consists of linear layers with LayerNorm and GELU activation. This projects from the LLM's hidden dimension ($d_{\text{llm}} = 2560$ for Qwen3-4B, $4096$ for Qwen3-8B) to the shared representation space ($d_p = 256$).

\subsection{Definition-Guided Entity Typing}
\label{sec:definitions}

A key component enabling zero-shot generalization is our use of natural language definitions to represent entity types. Rather than encoding entity types by their names alone (e.g., ``PERSON''), which relies on surface-level semantics, we encode rich definitions that specify exactly what should be tagged.

This approach enables flexible zero-shot transfer: new entity types can be specified at inference time without retraining, as the model learns to match token representations to definition semantics rather than memorizing fixed label vocabularies. Beyond generalization, definition-guided typing provides fine-grained control, allowing users to specify boundary cases directly in natural language. This makes JPT a controllable information extractor rather than a fixed NER model. We validate these properties in Section~\ref{sec:ablations}.

For each entity type $j$, we create a definition text $\text{def}_j$ and encode it using a pre-trained text embedding model:
\begin{equation}
\mathbf{p}_j = \text{MLP}_{\text{entity}}(\mathbf{enc}_j) \in \mathbb{R}^{d_p}
\end{equation}
where $\mathbf{enc}_j = \text{Embed}(\text{def}_j) \in \mathbb{R}^{d_{\text{enc}}}$ is the raw embedding from the text encoder (e.g., $d_{\text{enc}} = 4096$ for Qwen3-Embedding).

\paragraph{Definition Format.} Definitions can range from simple (``A human individual's name'') to detailed specifications that include boundary cases and disambiguation rules:
\begin{quote}
\small
\texttt{LOCATION}: ``Any word indicating WHERE: explicit place names (Boston, downtown), relative indicators (nearby, around), directional words (east, south side).''
\end{quote}

This flexibility allows users to precisely control what gets tagged. For instance, they can specify whether ``nearby'' should be tagged as a \texttt{LOCATION} indicator or ignored as a common adjective.

\paragraph{Dual-Channel Definition Injection.} In addition to encoding definitions as entity embeddings $\mathbf{p}_j$, we include the list of entity types and their definitions in the LLM's input prompt (see Appendix~\ref{sec:prompt}). This provides explicit semantic guidance during token encoding: the LLM's attention mechanism can attend to definitions while processing each token, while the embedding space enables the classifier to match tokens to entity semantics. Our ablation study (Section~\ref{sec:ablations}) shows that both channels provide complementary benefits.

\paragraph{Embedding Caching.} Entity type definitions remain constant at inference time, so their embeddings can be precomputed and cached, incurring no additional latency during prediction. Any text embedding model can be used, e.g., OpenAI's \texttt{text-embedding-3-small} (1536-dim) or Qwen3-Embedding (4096-dim).

\subsection{Classification}

Given projected token representations $\mathbf{t}_i \in \mathbb{R}^{d_p}$ and entity embeddings $\mathbf{P} = [\mathbf{p}_1, \ldots, \mathbf{p}_N] \in \mathbb{R}^{N \times d_p}$, we compute classification scores using a bilinear interaction:
\begin{equation}
s_{ij} = \mathbf{t}_i^\top \mathbf{W} \mathbf{p}_j + b_j
\label{eq:bilinear}
\end{equation}
where $\mathbf{W} \in \mathbb{R}^{d_p \times d_p}$ is a learned weight matrix and $b_j$ is a learned type-specific bias.

We include an explicit ``O'' (outside/non-entity) class with a fixed embedding $\mathbf{p}_O$ derived from the definition ``A token that is not part of any named entity,'' making this an $(N+1)$-way classification. The final prediction is:
\begin{equation}
\hat{y}_i = \arg\max_{j \in \{O, 1, \ldots, N\}} s_{ij}
\end{equation}

At inference, consecutive entity tokens are merged into spans, with span boundaries triggered by type changes.

\subsection{Training}
\label{sec:training}

\paragraph{Parameter Efficiency.} We freeze the LLM backbone and train only: (1) LoRA adapters \citep{hu2021loralowrankadaptationlarge} on the attention projections, (2) token and entity projection MLPs, and (3) the bilinear classifier. This yields strong performance while updating under 2\% of backbone parameters (0.95\% for JPT-4B, 1.71\% for JPT-8B). Full details in Appendix~\ref{sec:appendix-arch}.

\paragraph{Loss Function.} We use weighted classification losses (cross entropy and focal loss) to handle class imbalance, downweighting the ``O'' class relative to entity classes. Loss is computed only on tokens from the second pass. Full details on the loss functions are provided in Appendix~\ref{sec:appendix-arch}.

\paragraph{Training Data.} We train on an in-house NER dataset derived from Wikipedia with automatic annotation, containing diverse entity types. Importantly, this dataset has \emph{no overlap} with any evaluation benchmark, ensuring our evaluation is truly zero-shot with respect to test domains and datasets.

Full training data statistics and examples are provided in Appendix~\ref{sec:appendix-data}.

\section{Experiments}
\label{sec:experiments}

\subsection{Experimental Setup}

We evaluate on two established zero-shot NER benchmarks:
\begin{itemize}
    \item \textbf{CrossNER} \citep{liu2020crossner}: Five specialized domains (AI, Literature, Music, Politics, Science) with 9--17 domain-specific entity types per domain.
    \item \textbf{MIT Movie/Restaurant} \citep{6639301}: Slot-filling NER for conversational queries; the Restaurant dataset contains 8 entity types and the Movie dataset contains 12.
\end{itemize}

We compare against state-of-the-art methods from both paradigms:
\begin{itemize}
    \item \textit{Generative}: UniNER-7B \citep{zhou2024universalner}, GoLLIE \citep{sainz2024gollie}, InstructUIE \citep{wang2023instructuiemultitaskinstructiontuning}, GPT-NER \citep{wang-etal-2025-gpt}, SaM \citep{ding-etal-2025-selecting}
    \item \textit{Discriminative}: GLiNER-L \citep{zaratiana-etal-2024-gliner}
\end{itemize}

\paragraph{Implementation Details.} We use Qwen3-4B and Qwen3-8B as base LLMs. Entity embeddings use Qwen3-Embedding-8B. The shared projection dimension is $d_p=256$. Training uses AdamW with learning rate $5 \times 10^{-5}$, effective batch size 8, and 5 epochs on a 4xH100 GPU machine. LoRA and projection configurations are detailed in Section~\ref{sec:training}.

\subsection{Main Results}

\begin{table*}[t]
\centering
\small
\setlength{\tabcolsep}{4pt}
\begin{tabular}{lcccccccc}
\toprule
\textbf{Model} & \textbf{AI} & \textbf{Lit.} & \textbf{Music} & \textbf{Politics} & \textbf{Science} & \textbf{Movie} & \textbf{Rest.} & \textbf{Avg} \\
\midrule
\multicolumn{9}{l}{\textit{Generative Baselines}} \\
UniNER-7B & 62.9 & 64.9 & 70.6 & 66.9 & 70.8 & 61.2 & 35.2 & 61.8 \\
GoLLIE & 59.1 & 62.7 & 67.8 & 57.2 & 55.5 & 63.0 & 43.4 & 58.4 \\
InstructUIE & 49.0 & 47.2 & 53.2 & 48.2 & 49.3 & 63.0 & 21.0 & 47.8 \\
SaM (MoE) & 60.9 & 66.9 & 73.5 & 74.4 & 62.6 & 72.1 & 52.9 & 66.2 \\
\midrule
\multicolumn{9}{l}{\textit{Discriminative Baselines}} \\
GLiNER-L & 57.2 & 64.4 & 69.6 & 72.6 & 62.6 & 64.4 & 42.9 & 60.9 \\
\midrule
\textbf{JPT-4B (Ours)} & 68.3 & \textbf{73.7} & 84.1 & 76.4 & 69.5 & 60.7 & 63.4 & 70.9 \\
\textbf{JPT-8B (Ours)} & \textbf{71.9} & 72.2 & \textbf{85.3} & \textbf{77.0} & \textbf{71.3} & \textbf{76.5} & \textbf{64.4} & \textbf{74.1} \\
\bottomrule
\end{tabular}
\caption{Zero-shot F1 scores on CrossNER and MIT benchmarks. JPT-8B achieves the best overall average, improving by +7.9 F1 over the strongest baseline (SaM). Results for baselines are taken from prior work \citep{ding-etal-2025-selecting}.}
\label{tab:main_results}
\end{table*}

Table~\ref{tab:main_results} presents our main results. \textbf{JPT-8B} achieves state-of-the-art performance, outperforming the previous best method (\textbf{SaM}) by +7.9 F1 overall and (\textbf{UniNER-7B}) by +12.3 F1. The largest gains appear on Music (+11.8 F1), Restaurant (+11.5 F1), and AI (+11.0 F1), domains with specialized terminology where LLM world knowledge proves particularly beneficial.
Even \textbf{JPT-4B} surpasses all baselines (+4.7 F1 over SaM), with further gains from scaling to 8B, suggesting that larger LLM backbones continue to improve performance.

\subsection{Extended Benchmark Results}
\begin{table}[t]
\centering
\small
\setlength{\tabcolsep}{4pt}
\begin{tabular}{l|cc|c}
\hline
\textbf{Dataset} & \textbf{UniNER-7B} & \textbf{GLiNER-L} & \textbf{JPT-4B} \\
\hline\hline
ACE05 & 36.9 & 27.3 & \textbf{44.6} \\
AnatEM & 25.1 & 33.3 & \textbf{37.2} \\
bc2gm & 46.2 & 47.9 & \textbf{54.8} \\
bc4chemd & 47.9 & 43.1 & \textbf{53.3} \\
bc5cdr & 68.0 & 66.4 & \textbf{70.4} \\
Broad Twitter & 67.9 & 61.2 & \textbf{71.2} \\
CoNLL03 & 72.2 & 64.6 & \textbf{78.1} \\
FabNER & 24.8 & 23.6 & \textbf{27.8} \\
FindVehicle & 22.2 & 41.9 & \textbf{42.3} \\
GENIA & 54.1 & \textbf{55.5} & 50.8 \\
HarveyNER & 18.2 & 22.7 & \textbf{27.7} \\
MIT Movie & 42.4 & 57.2 & \textbf{73.4} \\
MIT Restaurant & 31.7 & 42.9 & \textbf{61.9} \\
MultiNERD & 59.3 & 59.7 & \textbf{65.7} \\
NCBI & 60.4 & 61.9 & \textbf{68.7} \\
OntoNotes & 27.8 & 32.2 & \textbf{43.1} \\
PolyglotNER & 41.8 & 42.9 & \textbf{47.4} \\
TweetNER7 & 42.7 & 41.4 & \textbf{49.7} \\
WikiANN & 55.4 & 58.9 & \textbf{64.7} \\
WikiNeural & 69.2 & 71.8 & \textbf{77.3} \\
\hline\hline
\textbf{Average} & 45.7 & 47.8 & \textbf{55.5} \\
\hline
\end{tabular}
\caption{Extended zero-shot F1 results across 20 NER benchmarks spanning biomedical, social media, and multilingual domains. Results for UniNER-7B and GLiNER-L are reported from \citep{zaratiana-etal-2024-gliner}.}
\label{tab:extended-results}
\end{table}

Table~\ref{tab:extended-results} reports results on an extended suite of 20 datasets. \textbf{JPT-4B} outperforms both \textbf{GLiNER-L} and \textbf{UniNER-7B} on 19 out of 20 benchmarks.

\subsection{Efficiency Analysis}
\label{sec:efficiency}

\begin{table}[t]
\centering
\small
\setlength{\tabcolsep}{3pt}
\begin{tabular}{l|c|c|c}
\hline
\textbf{Method} & \textbf{Cost(\$)$\downarrow$} & \textbf{Time (sec)$\downarrow$} & \textbf{F1$\uparrow$} \\
\hline\hline
\multicolumn{4}{l}{\textit{Generative Methods}} \\
\hline
UniNER-7B & 2.77 & 1970.2 & 61.8 \\
GNER & 8.21 & 5831.5 & 75.8 \\
GPT-5$^\dagger$ & 0.49 & 579.6 & 67.1 \\
\hline
\multicolumn{4}{l}{\textit{Discriminative Methods}} \\
\hline
GLiNER-L & \textbf{0.05} & \textbf{33.3} & 60.9 \\
\hline
\textbf{JPT-4B} & 0.13 & 89.7 & 76.4 \\
\textbf{JPT-8B} & 0.21 & 146.2 & \textbf{77.0} \\
\hline
\end{tabular}
\caption{Efficiency comparison on CrossNER-Politics. Cost: compute time $\times$ \$5.07/hour (local) or API pricing (GPT). $^\dagger$GPT results vary with prompting strategy.}
\label{tab:efficiency}
\end{table}

We estimate cost using inference time $\times$ \$5.07/hour for local models or API pricing for GPT-5, adopting the NER prompting template from \citet{ye2023comprehensivecapabilityanalysisgpt3}. All local models are benchmarked on the same A100 GPU with batch size 1 for fair comparison. Table~\ref{tab:efficiency} summarizes results.

\textbf{JPT} processes all tokens in a single forward pass, while generative methods must decode token-by-token, with latency scaling with output length. \textbf{JPT-4B} is \textbf{$\approx$22$\times$ faster} than \textbf{UniNER-7B}, and even \textbf{JPT-8B} (with a larger backbone and doubled input length) remains \textbf{$\approx$13.5$\times$ faster} while achieving higher F1.

\paragraph{Why Doubling Input Length is Fast.} While JPT doubles the input sequence length via concatenation (processing $2N$ tokens), this operation occurs entirely within the \emph{prefill phase}, which exploits massive parallelism and is compute-bound, achieving high GPU utilization \citep{wang2025systematiccharacterizationllminference}. In contrast, generative approaches rely on the \emph{decode phase}, which is inherently sequential and memory-bound; each generated token requires reloading model weights from memory for a single step of computation, leaving compute units underutilized \citep{patel2024splitwiseefficientgenerativellm}. Consequently, a single forward pass over a duplicated input is orders of magnitude faster than autoregressively generating entity lists, effectively trading cheap ``parallel'' input tokens for expensive ``serial'' output tokens. This efficiency gain is reflected in commercial API pricing, where providers typically charge 3--5$\times$ more for output tokens.

\section{Ablation Studies and Analysis}
\label{sec:analysis}

\subsection{Ablation Studies}
\label{sec:ablations}

We validate our two core design choices, input duplication and definition-guided typing, through ablation experiments. We fine-tune several JPT variants to isolate the impact of each design choice. These include: a single-input variant without duplication, a variant with definitions incorporated only in the input prompt, another with definitions integrated solely into the embedding model, and a baseline without definitions that relies only on type labels. Table~\ref{tab:ablations} summarizes results on CrossNER and MIT benchmarks.

\begin{table}[t]
\centering
\small
\setlength{\tabcolsep}{4pt}
\begin{tabular}{l|c}
\hline
\textbf{Configuration} & \textbf{Avg Micro F1} \\
\hline
\multicolumn{2}{l}{\textit{Input Duplication}} \\
\hline
Single Pass & 55.7 \\
\hline
\multicolumn{2}{l}{\textit{Entity Definitions}} \\
\hline
No Definitions & 58.3 \\
Prompt-only definitions & 63.3 \\
Embedding-only definitions & 65.2 \\
\hline
Dual-channel definitions + Double Pass (JPT) & \textbf{70.9} \\
\hline
\end{tabular}
\caption{Ablation studies on JPT-4B. Avg.\ micro-F1 across CrossNER and MIT benchmarks. Input duplication provides +15.2 F1; dual-channel definitions provide +12.6 F1 over no definitions.}
\label{tab:ablations}
\end{table}

\paragraph{Input Duplication.} Removing input duplication (single pass) degrades performance by $-$15.2 F1, confirming that bidirectional context is essential for effective token classification with causal LLMs.

\paragraph{Entity Definitions.} Using definitions in only one channel (prompt or embedding) provides limited gains. Jointly injecting definitions into both the prompt and embeddings yields the highest performance (+12.6 F1 over no definitions), suggesting that the two mechanisms provide complementary signals. The ability to precisely define entity boundaries proved especially valuable. For example, on MIT-Restaurant, specifying that \texttt{LOCATION} includes relative indicators like ``nearby'' improved type-specific F1 by +32.6 points (see Table~\ref{tab:definition-examples} in Appendix~\ref{sec:appendix-definitions}).

\subsection{Understanding the Attention Patterns}
\begin{figure}[t]
    \centering
    \includegraphics[width=\columnwidth]{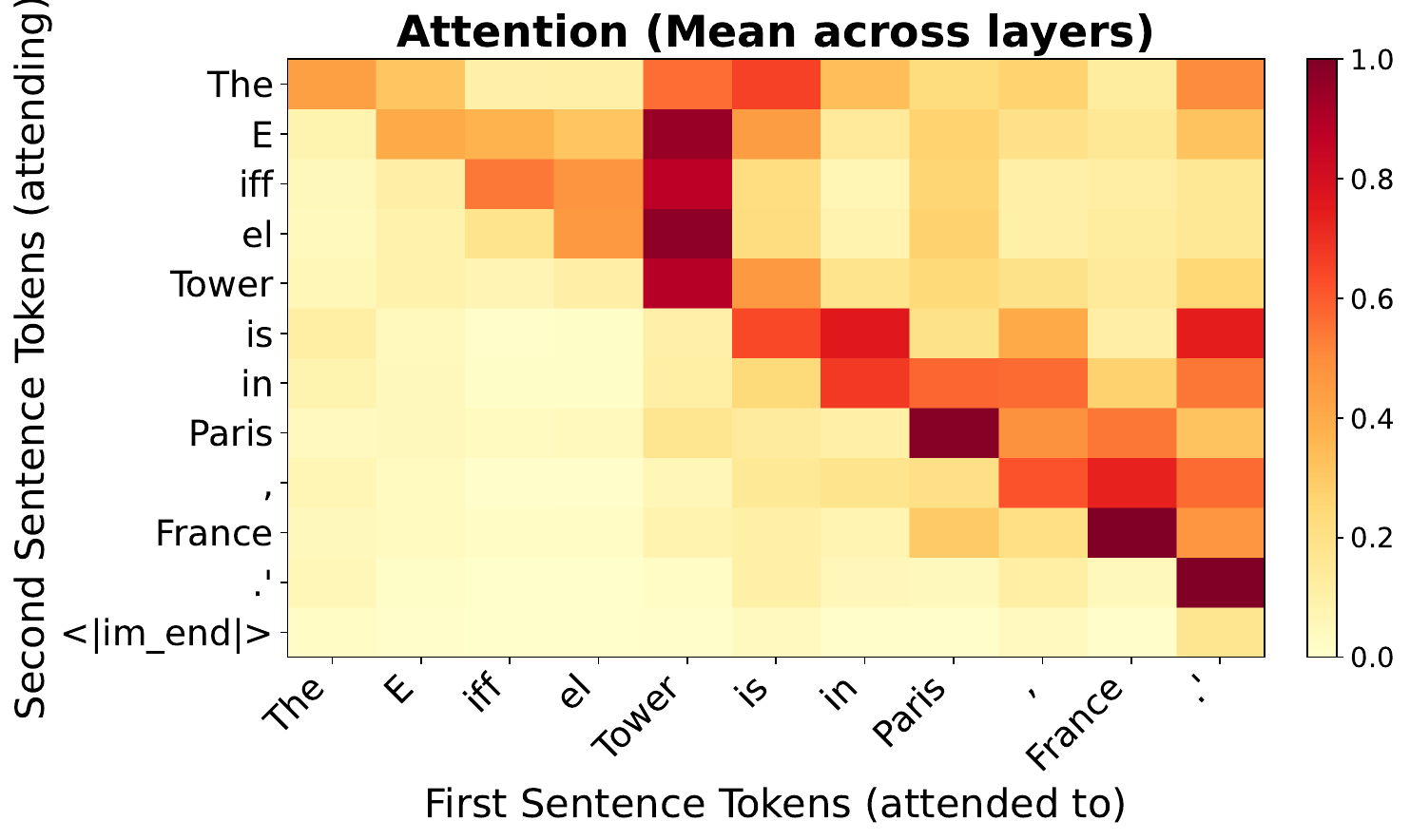}
    \vspace{-0.6em}
    \caption{Attention weights averaged across all transformer layers from second-pass tokens (rows) to first-pass tokens (columns). The model attends to corresponding positions (diagonal) and semantically relevant context, suggesting effective bidirectional information flow.}
    \label{fig:sentence_attention}
\end{figure}

The attention patterns reveal how JPT leverages bidirectional context for disambiguation. In Figure~\ref{fig:sentence_attention}, subword tokens ``The,'' ``E,'' ``iff,'' and ``el'' in the second pass attend most strongly to ``Tower'' and to the remaining subwords of their entity from the first pass. This context appears \emph{after} these tokens in the original sequence but becomes accessible through input duplication. This demonstrates JPT's core mechanism: incomplete tokens use the first pass to ``look ahead,'' attending to complete words and surrounding context that would otherwise be masked. This lookahead enables accurate entity boundary detection, correct type classification, and disambiguation of ambiguous mentions, allowing the model to form coherent entity representations despite the underlying causal constraint.

\subsection{Error Analysis}
A comprehensive error analysis including boundary detection errors, type confusion cases, and contrastive examples where SOTA baselines fail is provided in Appendix~\ref{sec:appendix-errors}.

\section{Conclusion}

We presented Just Pass Twice (JPT), a method enabling causal LLMs to perform discriminative token classification with bidirectional context via input duplication. Combined with definition-guided entity typing, JPT achieves state-of-the-art zero-shot NER results while being over 20$\times$ faster than generative alternatives. Our work demonstrates that causal attention constraints need not limit LLMs to generative approaches. The simplicity of our method suggests applicability to other token-level tasks beyond NER.

\section*{Limitations}

\begin{itemize}[leftmargin=*]
    \item \textbf{Sequence length:} Input duplication doubles the effective sequence length, increasing attention complexity from $O(N^2)$ to $O((2N)^2)$, which may pose memory constraints for long contexts. In practice, this is mitigated by chunking long documents into shorter segments that can be batched together, a standard practice that JPT supports efficiently since it operates entirely in the parallel prefill phase.
    
    \item \textbf{Flat NER only:} JPT currently assigns one label per token, so nested entities are not supported and span boundaries depend on post-hoc merging of consecutive predictions. However, the architecture could be extended to support nested entities by using the sigmoid head's independent per-type probabilities (predicting multiple types with probability $>0.5$ for overlapping spans).
    
    \item \textbf{Training data:} Unlike prior zero-shot NER methods that train on existing benchmark training splits, we use an entirely separate Wikipedia-derived dataset with no overlap with any evaluation benchmark. While common entity types (e.g., PERSON, LOCATION) appear in our training data with different definitions, we validated broad generalization across 20 diverse benchmarks spanning biomedical, social media, and multilingual domains.
\end{itemize}

\bibliography{custom}

\appendix

\section{Model Architecture and Training Details}
\label{sec:appendix-arch}

This section provides complete details on the JPT architecture, training configuration, and hyperparameters.

\subsection{Architecture Configuration}

Table~\ref{tab:arch-details} summarizes the architecture for both model variants. The base LLMs are frozen, with only lightweight adapters and projection layers trained.

\begin{table*}[t]
\centering
\small
\setlength{\tabcolsep}{6pt}
\renewcommand{\arraystretch}{1.1}
\begin{tabular}{l|c|c}
\toprule
\textbf{Component} & \textbf{JPT-4B} & \textbf{JPT-8B} \\
\midrule
\multicolumn{3}{l}{\textit{Base LLM (Frozen)}} \\
\midrule
Model & Qwen3-4B & Qwen3-8B \\
Hidden dim ($d_{\text{llm}}$) & 2560 & 4096 \\
Layers / Attention heads & 36 / 32 & 36 / 32 \\
\midrule
\multicolumn{3}{l}{\textit{LoRA Adapters (Trained)}} \\
\midrule
Rank ($r$) / Alpha ($\alpha$) & 32 / 64 & 128 / 256 \\
Target modules & \multicolumn{2}{c}{\texttt{q\_proj}, \texttt{k\_proj}, \texttt{v\_proj}, \texttt{o\_proj}} \\
Parameters & $\sim$23.6M & $\sim$122.7M \\
\midrule
\multicolumn{3}{l}{\textit{Projection MLPs (Trained)}} \\
\midrule
Token MLP & $d_{\text{llm}}\!\rightarrow\!1024\!\rightarrow\!d_p$ & $d_{\text{llm}}\!\rightarrow\!1024\!\rightarrow\!512\!\rightarrow\!d_p$ \\
Entity MLP & $d_{\text{enc}}\!\rightarrow\!1024\!\rightarrow\!d_p$ & $d_{\text{enc}}\!\rightarrow\!1024\!\rightarrow\!512\!\rightarrow\!d_p$ \\
Shared dim ($d_p$) & 256 & 256 \\
Parameters & $\sim$14.7M & $\sim$19.4M \\
\midrule
\multicolumn{3}{l}{\textit{Classifier (Trained)}} \\
\midrule
Dual bilinear classifiers & $\sim$131K & $\sim$131K \\
\midrule
\textbf{Total trainable} & \textbf{$\sim$38.8M (0.95\%)} & \textbf{$\sim$142.8M (1.71\%)} \\
\bottomrule
\end{tabular}
\caption{Detailed architecture configuration for JPT models. Only a small fraction of backbone parameters are trained via LoRA adapters and lightweight projection heads.}
\label{tab:arch-details}
\end{table*}

\subsection{Classifier and Loss Details}
While Section~\ref{sec:method} describes classification using a single bilinear scorer for simplicity, our implementation uses an ensemble of two classifiers for improved calibration:

\begin{itemize}[nosep]
   
   \item \textbf{Softmax head}: Models mutually exclusive token labels (including an explicit O-class). Trained with cross-entropy loss, where the O-class weight is reduced to $w_O = 0.25$ to handle class imbalance.
   \item \textbf{Sigmoid head}: Treats each entity type as an independent binary decision. Trained with focal loss ($\gamma = 2.5$, positive weight $= 5.0$) to focus on hard examples and upweight rare 
    entities.
\end{itemize}

The two heads share the same projected token and entity embeddings. Final predictions are obtained by averaging their probability outputs. This ensemble provides complementary strengths: the softmax head yields decisive predictions while the sigmoid head better handles rare entities. The core method (input duplication + definition-guided typing) is independent of this design choice; the ensemble improves F1 by 1--2 points over a single softmax head.

\subsection{Training Hyperparameters}
Table~\ref{tab:training-hyperparams} summarizes the training hyperparameters used for both \textbf{JPT-4B} and \textbf{JPT-8B}.

\begin{table}[h]
\centering
\small
\begin{tabular}{l|c}
\toprule
\textbf{Hyperparameter} & \textbf{Value} \\
\midrule
Optimizer & AdamW \\
Learning rate & $5 \times 10^{-5}$ \\
LR scheduler & Cosine with warmup \\
Warmup ratio & 10\% of total steps \\
Effective batch size & 8 \\
Gradient accumulation & 2 \\
Training epochs & 5 \\
Max sequence length & 4096 \\
O-class weight ($w_O$) & 0.25 \\
Entity class weight & 1.0 \\
Hardware & 4$\times$ H100 GPU \\
Training time & $\sim$1.5 hours \\
\bottomrule
\end{tabular}
\caption{Training hyperparameters used for both JPT-4B and JPT-8B.}
\label{tab:training-hyperparams}
\end{table}

\section{Training Data}
\label{sec:appendix-data}
This section describes the training corpus used to train \textbf{JPT}, including dataset statistics, construction procedure, entity-type distribution, and representative annotated examples.

\subsection{Dataset Statistics}
Table~\ref{tab:training-stats} reports the main training dataset statistics, including corpus size, token counts, and entity-type diversity.

\begin{table}[h]
\centering
\small
\begin{tabular}{l|r}
\toprule
\textbf{Statistic} & \textbf{Value} \\
\midrule
Total sentences & 17,489 \\
Total tokens & 3,391,899 \\
Total entity mentions & 374,705 \\
Unique entity types & 5,009 \\
Avg. sentence length & 194.9 tokens \\
Entity/non-entity ratio & 1:1.5 \\
\bottomrule
\end{tabular}
\caption{Training dataset statistics.}
\label{tab:training-stats}
\end{table}

\subsection{Dataset Construction}
Our training data consists of \emph{natural text} from Wikipedia articles, sourced from the test partition of the DBpedia corpus in the TELEClass benchmark \cite{zhang2025teleclass}. Each passage is associated with a three-level hierarchical topic taxonomy from the DBpedia ontology, progressing from broad categories (e.g., ``agent,'' ``place'') through intermediate concepts (e.g., ``athlete,'' ``natural\_place'') to fine-grained types (e.g., ``chess\_player,'' ``mountain'')

We use Claude Sonnet 4.5 (with extended thinking) to automatically annotate these real passages in two stages:
\begin{enumerate}[nosep]
    \item \textbf{Type Generation}: Given a passage and its topic hierarchy, the model proposes domain-appropriate entity types with definitions (e.g., ``Athlete,'' ``Team,'' ``Stadium'' for sports articles).
    \item \textbf{Entity Detection}: The model identifies entity spans in the text, followed by a gap-detection pass to catch missed mentions.
\end{enumerate}

For quality validation, Claude Opus 4.5 assesses random samples across entity type appropriateness, definition actionability, and extraction accuracy. The resulting dataset comprises \textbf{17,489} training examples with \textbf{5,009} entity types and \textbf{2,500} test examples with \textbf{1,947} entity types.

\subsection{Entity Type Distribution}
\begin{figure}[t]
\centering
\includegraphics[width=\columnwidth]{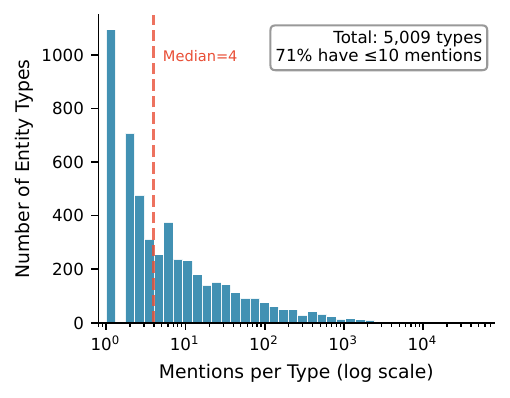}
\caption{Distribution of entity type frequencies in training data. The long-tail distribution (71\% of types have $\leq$10 mentions, median=4) encourages the model to leverage definition semantics rather than memorizing patterns.}
\label{fig:entity-dist}
\end{figure}

Figure~\ref{fig:entity-dist} shows a long-tailed entity type distribution, with most types occurring infrequently, encouraging reliance on definition-based generalization rather than frequency-driven learning.

\subsection{Training Examples}

Figure~\ref{fig:training-examples} shows representative training examples with fine-grained entity annotations.

\begin{figure*}[t]
\centering
\small
\setlength{\tabcolsep}{8pt}
\renewcommand{\arraystretch}{1.4}
\begin{tabular}{@{}p{0.97\textwidth}@{}}
\toprule
\textbf{Example 1: Sports Domain} \\
\midrule
``the \textcolor{entitycolor}{\underline{men's foil competition}}$_{\textsc{\scriptsize[Competition]}}$ in \textcolor{entitycolor}{\underline{fencing}}$_{\textsc{\scriptsize[Sport]}}$ at the \textcolor{entitycolor}{\underline{2016 olympic games}}$_{\textsc{\scriptsize[Event]}}$ in \textcolor{entitycolor}{\underline{rio de janeiro}}$_{\textsc{\scriptsize[City]}}$ was held on \textcolor{entitycolor}{\underline{7 august}}$_{\textsc{\scriptsize[Date]}}$ at the \textcolor{entitycolor}{\underline{carioca arena 3}}$_{\textsc{\scriptsize[Venue]}}$. the medals were presented by \textcolor{entitycolor}{\underline{paul tergat}}$_{\textsc{\scriptsize[Person]}}$, \textcolor{entitycolor}{\underline{ioc}}$_{\textsc{\scriptsize[Org]}}$ member, \textcolor{entitycolor}{\underline{kenya}}$_{\textsc{\scriptsize[Country]}}$'' \\
\midrule
\textbf{Example 2: Natural Disaster Domain} \\
\midrule
``the \textcolor{entitycolor}{\underline{2014 mae lao earthquake}}$_{\textsc{\scriptsize[Earthquake]}}$ occurred at \textcolor{entitycolor}{\underline{18:08:43}}$_{\textsc{\scriptsize[Time]}}$ \textcolor{entitycolor}{\underline{indochina time}}$_{\textsc{\scriptsize[TimeZone]}}$ on \textcolor{entitycolor}{\underline{may 5}}$_{\textsc{\scriptsize[Date]}}$. the epicenter was located \textcolor{entitycolor}{\underline{9 km}}$_{\textsc{\scriptsize[Distance]}}$ south of \textcolor{entitycolor}{\underline{mae lao district}}$_{\textsc{\scriptsize[District]}}$, \textcolor{entitycolor}{\underline{27 km}}$_{\textsc{\scriptsize[Distance]}}$ southwest of \textcolor{entitycolor}{\underline{chiang rai}}$_{\textsc{\scriptsize[City]}}$, \textcolor{entitycolor}{\underline{thailand}}$_{\textsc{\scriptsize[Country]}}$'' \\
\midrule
\textbf{Example 3: Biographical Domain} \\
\midrule
``\textcolor{entitycolor}{\underline{helga mühlberg-ulze}}$_{\textsc{\scriptsize[Athlete]}}$ is an \textcolor{entitycolor}{\underline{east german}}$_{\textsc{\scriptsize[Nationality]}}$ \textcolor{entitycolor}{\underline{sprint canoeist}}$_{\textsc{\scriptsize[SportDisc]}}$ who competed in the \textcolor{entitycolor}{\underline{early to mid 1960s}}$_{\textsc{\scriptsize[TimePeriod]}}$. she won \textcolor{entitycolor}{\underline{gold}}$_{\textsc{\scriptsize[Medal]}}$ (\textcolor{entitycolor}{\underline{k-2 500m}}$_{\textsc{\scriptsize[RaceCat]}}$: \textcolor{entitycolor}{\underline{1966}}$_{\textsc{\scriptsize[Year]}}$) and two \textcolor{entitycolor}{\underline{bronzes}}$_{\textsc{\scriptsize[Medal]}}$'' \\
\bottomrule
\end{tabular}
\caption{Training examples showing fine-grained entity types across diverse domains. Entity spans are \textcolor{entitycolor}{\underline{underlined in blue}} with type labels in subscript. The dataset contains 5,009 unique entity types including domain-specific categories like \textsc{Earthquake}, \textsc{TimeZone}, and \textsc{RaceCategory}.}
\label{fig:training-examples}
\end{figure*}

\section{Prompt Template}
\label{sec:prompt}

Figure~\ref{fig:prompt-template} shows the prompt structure used during training and inference. Definitions are injected in the user turn, providing the dual-channel signal described in Section~\ref{sec:definitions}.

\begin{figure*}[t]
\centering
\fbox{\parbox{0.95\textwidth}{
\ttfamily\small

\textcolor{blue!70!black}{\textbf{<|im\_start|>system}}\\[0.2em]
You are an information-extraction assistant.\\
Task: Perform Named Entity Recognition (NER) on the user-supplied text.\\
The user will give you the supported entity types and their definitions.\\
You will read the types and definitions to understand what each entity type means.\\
The user will give you the text twice in the format "The first time: 'actual text' The second time: 'actual text'".\\[0.2em]
\textbf{Output Format:} Output ONE annotated text with entities as \texttt{<entity\_text, ENTITY\_TYPE>}\\[0.2em]
\textbf{Rules:} \\
(1) Keep multi-word entities together; \\
(2) Only use provided types; \\
(3) Output once;\\
(4) No bare-noun labelling (e.g., don't label "museum" unless part of proper name);\\
(5) Output types exactly as listed; \\
(6) Only label if clearly matches definition.\\
\textcolor{blue!70!black}{\textbf{<|im\_end|>}}\\[0.4em]
\hrule\vspace{0.4em}

\textcolor{green!50!black}{\textbf{<|im\_start|>user}}\\[0.2em]
\textbf{Supported entity types (3):} ["PERSON", "ORGANIZATION", "LOCATION"]\\
\textbf{Entity type definitions:}\\
- "PERSON": "A named individual, including fictional characters"\\
- "ORGANIZATION": "A company, institution, or group with a formal name"\\
- "LOCATION": "A geographical place such as a city, country, or landmark"\\
\textcolor{green!50!black}{\textbf{<|im\_end|>}}\\[0.4em]
\hrule\vspace{0.4em}

\textcolor{purple!70!black}{\textbf{<|im\_start|>assistant}}\\[0.2em]
I have read the definitions. Please provide the text in the format 'The first time: <text> The second time: <text>'\\
\textcolor{purple!70!black}{\textbf{<|im\_end|>}}\\[0.4em]
\hrule\vspace{0.4em}

\textcolor{green!50!black}{\textbf{<|im\_start|>user}}\\[0.2em]
The first time: '<Input\_Sequence>' The second time: '<Input\_Sequence>'\\
\textcolor{green!50!black}{\textbf{<|im\_end|>}}
}}
\caption{Complete prompt template for \textbf{JPT}. The system prompt specifies output format and labeling rules. Entity definitions are injected in the first user turn, and the input text is duplicated with explicit markers in the second user turn.}
\label{fig:prompt-template}
\end{figure*}

\section{Definition Engineering}
\label{sec:appendix-definitions}

This section provides guidance on crafting effective entity definitions and demonstrates their impact on recognition quality.

\subsection{Impact of Definition Quality}

Table~\ref{tab:definition-examples} compares generic versus precise definitions. Precise definitions that specify boundary cases and provide examples yield substantial improvements.

\begin{table*}[t]
\centering
\small
\setlength{\tabcolsep}{4pt}
\begin{tabular}{p{1.2cm}|p{4.5cm}|p{6.5cm}|c}
\toprule
\textbf{Type} & \textbf{Generic Definition} & \textbf{Precise Definition} & \textbf{$\Delta$F1} \\
\midrule
\textsc{Location} & ``A geographical place'' & ``Any word indicating WHERE: explicit places (NYC, downtown), relative indicators (nearby, around, close by), directional phrases (east, south side). Tag the location word itself.'' & \textbf{+32.6} \\
\midrule
\textsc{Price} & ``A monetary value'' & ``Explicit monetary amounts (\$50, 100 dollars) AND qualitative price indicators (cheap, expensive, budget-friendly, overpriced, pricey).'' & \textbf{+34.9} \\
\midrule
\textsc{Amenity} & ``An available service'' & ``A feature, facility, or service offered by a restaurant. Includes: physical features (bar, parking), services (delivery, reservations, takeout), atmosphere descriptors (romantic, casual, family-friendly).'' & \textbf{+14.29} \\
\bottomrule
\end{tabular}
\caption{Generic vs. precise entity definitions and their impact. F1 computed per-type on MIT Restaurant. Precise definitions with boundary cases and examples dramatically improve recognition.}
\label{tab:definition-examples}
\end{table*}

\subsection{Definition Writing Guidelines}

Effective definitions should:
\begin{enumerate}[nosep]
    \item \textbf{Specify inclusions and exclusions}: What counts and what doesn't
    \item \textbf{Provide concrete examples}: Representative instances of the type
    \item \textbf{Address ambiguities}: Clarify edge cases (e.g., does ``nearby'' count as location?)
\end{enumerate}

\section{Additional Ablations}
\label{sec:additional-ablations}
This section presents additional ablation studies that analyze the effect of model scale and parameter-efficient adaptation on \textbf{JPT} performance. These experiments help characterize the trade-offs between accuracy, model capacity, and inference efficiency.

\subsection{Impact of LLM Size}
Table~\ref{tab:model-size} examines the impact of the base LLM size on performance and inference latency. Increasing model size consistently improves token-level F1, but the gains diminish as scale grows. This highlights a practical trade-off between accuracy and efficiency, motivating the use of mid-sized backbones in resource-constrained settings.

\begin{table}[h]
\centering
\small
\begin{tabular}{l|c|c}
\toprule
\textbf{Model} & \textbf{Params} & \textbf{Token F1} \\
\midrule
Qwen3-1.7B & 1.7B & 92.4 \\
Qwen3-4B & 4B & 93.9 \\
Qwen3-8B & 8B & 94.1 \\
Qwen3-14B & 14B & 94.5 \\
\bottomrule
\end{tabular}
\caption{Impact of base LLM size on token-level F1 (private evaluation set). Larger models yield consistent improvements, with diminishing returns beyond 8B parameters.}
\label{tab:model-size}
\end{table}

\subsection{Impact of LoRA Rank}
Table~\ref{tab:lora-rank} analyzes the effect of the LoRA rank on performance. Increasing the rank improves token-level F1, particularly when moving from frozen backbones to low-rank adaptation, but yields diminishing returns at higher ranks.

\begin{table}[h]
\centering
\small
\begin{tabular}{l|c}
\toprule
\textbf{LoRA Rank} & \textbf{Token F1} \\
\midrule
$r=0$ (frozen) & 82.0 \\
$r=16$ & 93.5 \\
$r=32$ (default) & 93.9 \\
$r=64$ & 94.2 \\
$r=128$ & 94.5 \\
$r=256$ & 94.7 \\
\bottomrule
\end{tabular}
\caption{Impact of LoRA rank on token-level F1 using JPT-4B (private evaluation set). Adaptation is essential ($r=0$ drops 12 points), but returns diminish beyond $r=32$.}
\label{tab:lora-rank}
\end{table}

\subsection{Comparison with Causal Backbone}
\label{sec:causal-backbone}
To isolate the contribution of the proposed framework from the underlying language model, we compare \textbf{JPT-4B} against its backbone, the causal \textit{Qwen3-4B} model. In this experiment, the base model is used in a purely generative setting. We adopt the NER prompting template from \citet{ye2023comprehensivecapabilityanalysisgpt3}.

Table~\ref{tab:causal-backbone} shows that \textbf{JPT-4B} significantly outperforms its backbone across all domains, achieving an average improvement of \textbf{+17.0 F1}. This result demonstrates that the performance gains are primarily driven by the proposed framework, namely input duplication, definition-guided typing, and discriminative classification, rather than the inherent capabilities of the underlying LLM.

In addition to accuracy improvements, we observe that the generative baseline is considerably slower at inference time due to autoregressive decoding, whereas JPT benefits from a more efficient discriminative formulation. These findings further highlight the practical advantages of the proposed approach in real-world deployment settings.

\begin{table}[t]
\centering
\small
\resizebox{\columnwidth}{!}{%
\begin{tabular}{lcccccc}
\toprule
\textbf{Model} & \textbf{AI} & \textbf{Lit.} & \textbf{Music} & \textbf{Politics} & \textbf{Science} & \textbf{Avg.} \\
\midrule
Qwen3-4B & 49.9 & 47.9 & 74.7 & 56.2 & 58.1 & 57.4 \\
\textbf{JPT-4B} & \textbf{68.3} & \textbf{73.7} & \textbf{84.1} & \textbf{76.4} & \textbf{69.5} & \textbf{74.4} \\
\bottomrule
\end{tabular}%
}
\caption{Comparison between \textbf{JPT-4B} and its causal backbone, \textit{Qwen3-4B}, on the CrossNER benchmark. JPT-4B consistently outperforms the backbone across all domains.}
\label{tab:causal-backbone}
\end{table}

\subsection{Why Single-Input Models Struggle Without Future Context}
\label{sec:single-input-future-context}

To better understand the performance gap between the single-input and dual-input settings, we analyze representative errors made by the single-input variant. Importantly, this model is not obtained by disabling duplication at inference time; rather, it is a dedicated \textbf{JPT} variant trained specifically with a single input as part of the ablation study. This ensures that the comparison reflects a fair architectural ablation rather than a mismatch between training and inference.

Our analysis shows that single-input models often struggle when the entity appears early in the sentence and its correct type can only be resolved using information that appears later. In such cases, the model must make a prediction without access to the future contextual cues that are necessary for disambiguation. These errors commonly manifest as type confusions or completely missed entities.

Table~\ref{tab:single-input-future-context} presents illustrative examples. In the first group, the single-input model predicts an entity span but assigns the wrong type because the discriminative evidence appears later in the sentence. In the second group, the model fails to predict any entity at all, again because decisive contextual information is only available in the following words. These examples support our core hypothesis that future context plays an important role in resolving fine-grained entity semantics.

\begin{table*}[t]
\centering
\small
\begin{tabular}{p{5.1cm}p{1.5cm}p{2.7cm}p{6.3cm}}
\toprule
\textbf{Entity at start of sentence} & \textbf{GT Type} & \textbf{Single-input prediction} & \textbf{Why future context matters} \\
\midrule
\multicolumn{4}{l}{\textbf{Type confusion at sentence start}} \\
\midrule
\textit{"Autoencoders are trained to minimise..."} & algorithm & product & Need ``trained to minimise'' to determine that it refers to an algorithm \\
\textit{"Fran\c{c}ois Bayrou (...) is a French centrist politician..."} & politician & person & Need ``politician'' later in the sentence to resolve the fine-grained type \\
\midrule
\multicolumn{4}{l}{\textbf{Completely missed at sentence start}} \\
\midrule
\textit{"The Color Purple is a 1982 epistolary novel..."} & book & (missed) & Need ``novel'' to recognize it as a book title \\
\textit{"Young's work on squid giant axons..."} & scientist & (missed) & Need scientific context later in the sentence to identify the mention as a scientist \\
\bottomrule
\end{tabular}
\caption{Representative errors made by the single-input \textbf{JPT} variant. When entity mentions appear early in the sentence, the absence of future context can lead to type confusion or missed predictions.}
\label{tab:single-input-future-context}
\end{table*}

\section{Attention Visualization}
\label{sec:appendix-attention}

\begin{figure*}[t]
\centering
\includegraphics[width=0.45\textwidth]{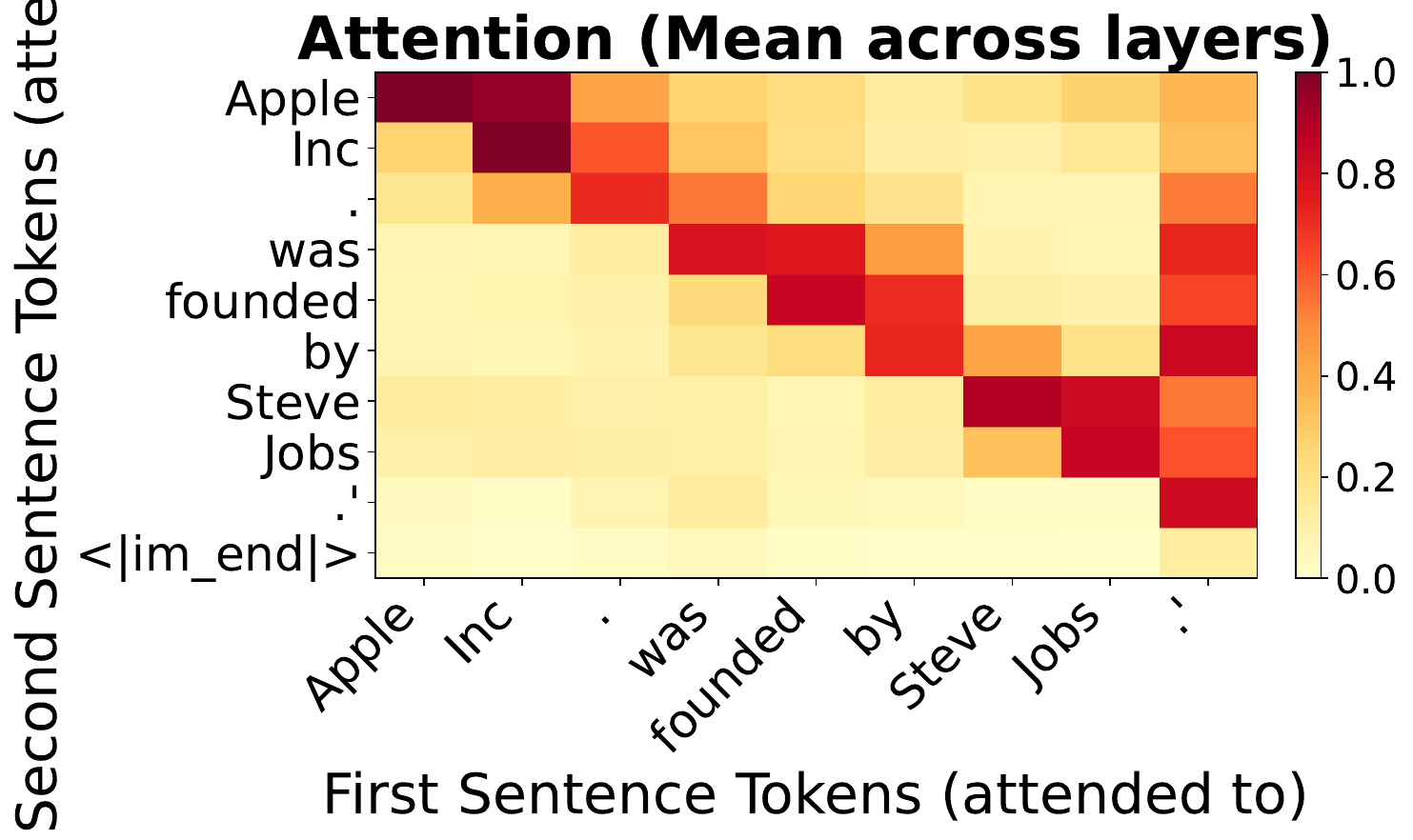}
\hfill
\includegraphics[width=0.45\textwidth]{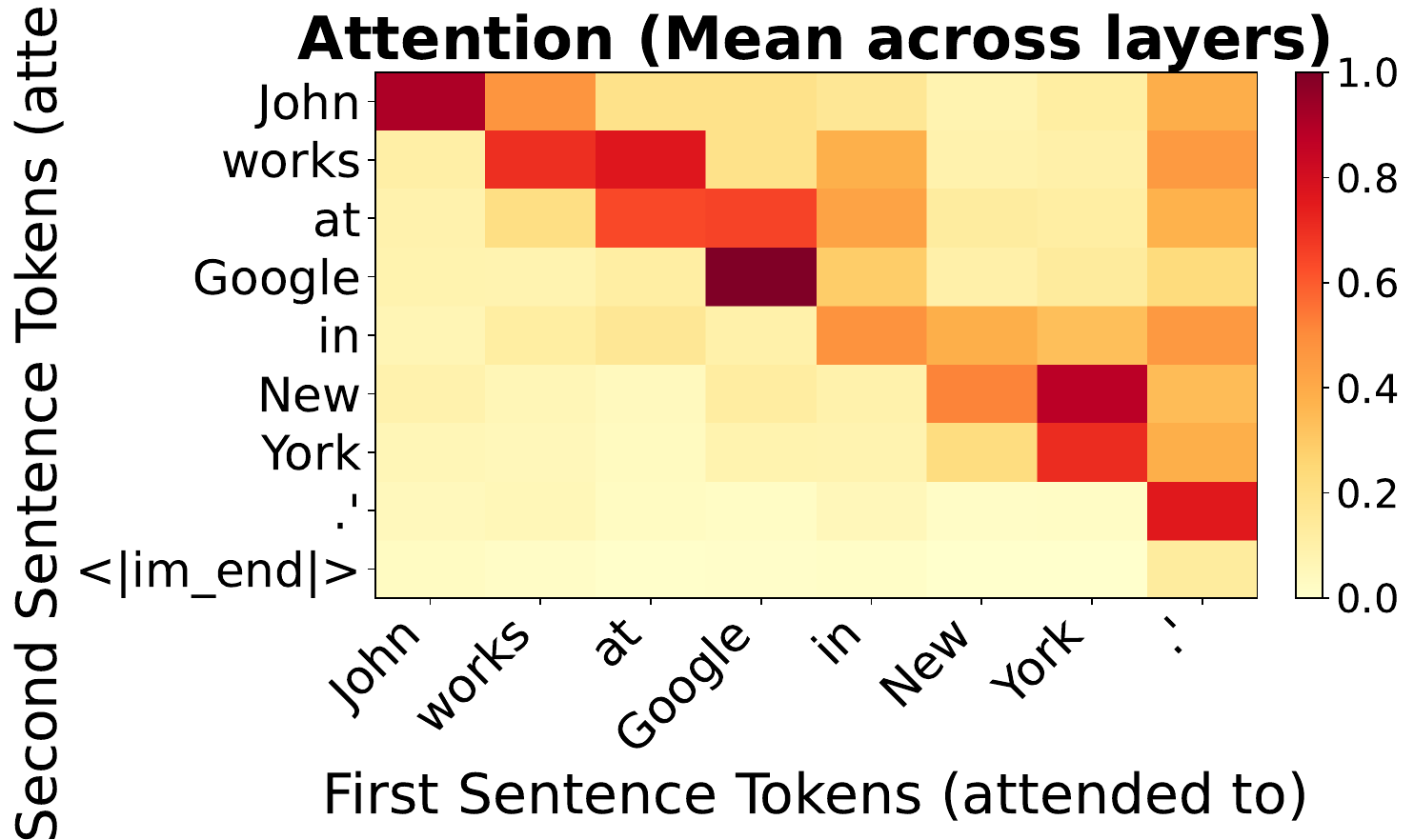}
\caption{Additional attention visualizations. Rows: second-pass tokens. Columns: first-pass tokens. The diagonal pattern (self-attention across passes) combined with off-diagonal attention to context tokens demonstrates effective bidirectional information flow.}
\label{fig:attention-examples}
\end{figure*}

Figure~\ref{fig:attention-examples} provides additional attention heatmaps illustrating bidirectional information flow across different sentence structures.

\section{Error Analysis}
\label{sec:appendix-errors}

We analyze common failure modes of JPT on the CrossNER and MIT benchmarks as well as custom examples.

\subsection{Boundary Detection Errors}
We observe that the most common failure mode of \textbf{JPT} involves boundary detection, where predicted spans partially overlap with the gold entities (Table~\ref{tab:boundary-errors}). These errors typically manifest as over-extension or truncation of entity boundaries, often due to nearby descriptive modifiers or appositional phrases.

\begin{table*}[t]
\centering
\small
\setlength{\tabcolsep}{6pt}
\begin{tabular}{l|l|l|l}
\hline
\textbf{Dataset} & \textbf{Entity Type} & \textbf{Gold Span} & \textbf{Predicted Span} \\
\hline
CrossNER-Science & ORGANIZATION & Virgo interferometer & Virgo interferometer collaboration \\
CrossNER-Science & ACADEMIC JOURNAL & the Springer Series Complexity & of the Springer Series Complexity \\
CrossNER-Politics & POLITICAL PARTY & Social Credit Party of Canada & national Social Credit Party of \\
CrossNER-Music & BAND & Collective Consciousness Society & The Collective Consciousness \\
CrossNER-Literature & LITERARY GENRE & Germany Romanticism & Romanticism era \\
MIT-Movie & TITLE & the reflecting skin & reflecting skin movie \\
MIT-Restaurant & LOCATION & within a mile & a mile of here \\
\hline
\end{tabular}
\caption{Representative boundary detection errors across CrossNER and MIT. JPT often predicts the correct type and core mention but may over-extend or truncate spans when entities appear with modifiers, appositions, or colloquial phrasing.}
\label{tab:boundary-errors}
\end{table*}

\subsection{Entity Type Confusion}
\label{sec:appendix-type-confusion}

\begin{figure*}[t]
\centering
\includegraphics[width=0.8\textwidth]{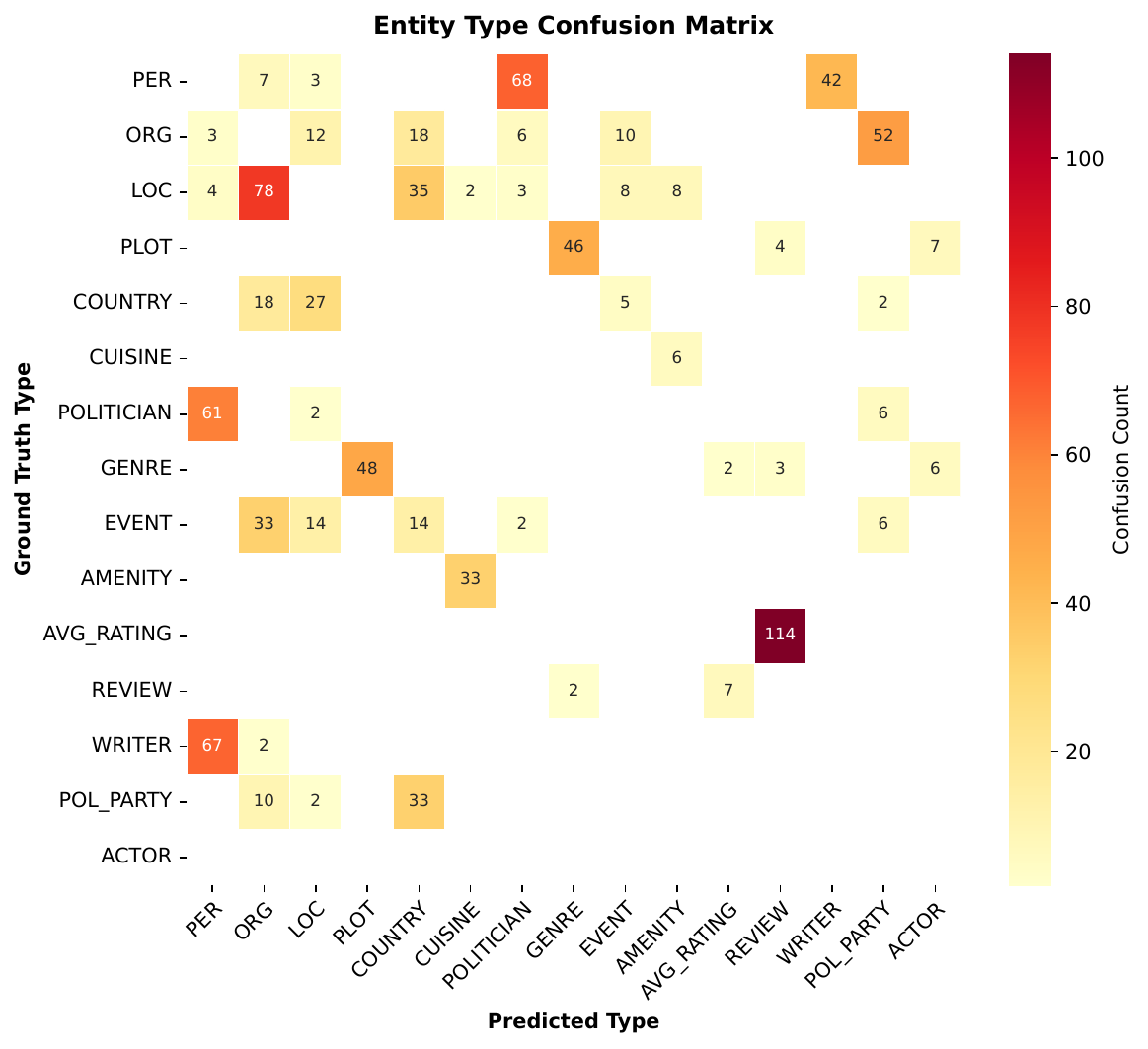}
\caption{Entity type confusion matrix aggregated across CrossNER and MIT. Most confusions occur between semantically adjacent types (e.g., \textsc{PER} vs.\ \textsc{ORG}, \textsc{LOC} vs.\ \textsc{COUNTRY}, and \textsc{POL\_PARTY} vs.\ \textsc{POLITICIAN}), suggesting that errors often stem from fine-grained boundary cases and overlapping semantic definitions rather than arbitrary label flips.}
\label{fig:type-confusion}
\end{figure*}

Figure~\ref{fig:type-confusion} shows the entity-type confusion matrix, highlighting systematic confusions between semantically related categories.

\subsection{Failure Examples}

Table~\ref{tab:failure-examples} presents representative failure cases from \textbf{JPT-4B} across diverse benchmarks. These examples illustrate three systematic error patterns: type confusion between semantically related categories, missed entities in atypical contexts, and over-predicted entities where plausible mentions lack ground-truth annotations.

These failure modes suggest that errors primarily stem from surface-form ambiguity and fine-grained semantic overlap rather than lack of contextual understanding. Type confusions often occur between closely related categories with overlapping definitions, missed entities frequently appear in descriptive or adjectival forms, and over-predictions arise when domain terms resemble entity mentions. Refining type definitions and adding targeted training examples for ambiguous and rare constructions may reduce these errors.

\subsection{SOTA Failures}
\label{sec:appendix-sota-failures}

Figure~\ref{fig:disambiguation-examples} provides contrastive disambiguation cases in which \textbf{JPT-4B} correctly predicts all entity spans and types, while \textbf{GLiNER} and \textbf{UniNER} exhibit errors. These examples highlight the benefit of definition-guided modeling for resolving polysemy in context.

\begin{figure*}[t]
\centering
\small
\fbox{\parbox{0.95\textwidth}{
\textbf{Example 1: Dutch Universities} (organization - completely missed)\\[0.5em]
\textit{``This school initially consisted of nearly 200 faculty members and Ph.D. students from the \underline{Vrije Universiteit}, \underline{University of Amsterdam}, \underline{Delft University of Technology}, and \underline{Leiden University}.''}\\[0.5em]
\textbf{Ground Truth:} Vrije Universiteit, University of Amsterdam, Delft University of Technology, Leiden University $\rightarrow$ ORGANIZATION\\[0.3em]
\textbf{Model Errors:}\\
\quad GLiNER: Missed all four universities entirely\\
\quad UniNER: Missed all four universities entirely\\
\quad JPT: Predicted all entities correctly
}}

\vspace{0.5em}

\fbox{\parbox{0.95\textwidth}{
\textbf{Example 2: Canadian Political Parties} (political party vs organization)\\[0.5em]
\textit{``In the 2006 Canadian federal election in Canada, the \underline{Liberal Party of Canada} used attack ads against \underline{Conservative Party of Canada} leader Stephen Harper.''}\\[0.5em]
\textbf{Ground Truth:} Liberal Party of Canada, Conservative Party of Canada $\rightarrow$ POLITICAL PARTY\\[0.3em]
\textbf{Model Errors:}\\
\quad GLiNER: Missed both political parties entirely\\
\quad UniNER: Over-predicted types for both parties (includes ORGANIZATION in addition to POLITICAL PARTY)\\
\quad JPT: Predicted all entities correctly
}}

\vspace{0.5em}

\fbox{\parbox{0.95\textwidth}{
\textbf{Example 3: US Political Leaders} (politician vs person)\\[0.5em]
\textit{``\underline{Lincoln} replaced \underline{Buell} with \underline{William Rosecrans}; and after the 1862 and 1863 United States House of Representatives elections he replaced \underline{McClellan} with \underline{Ambrose Burnside}.''}\\[0.5em]
\textbf{Ground Truth:} Lincoln, Buell, William Rosecrans, McClellan, Ambrose Burnside $\rightarrow$ POLITICIAN\\[0.3em]
\textbf{Model Errors:}\\
\quad GLiNER: Labeled all five as PERSON instead of POLITICIAN\\
\quad UniNER: Mixed predictions per name due to multi-type outputs; includes PERSON in addition to POLITICIAN\\
\quad JPT: Predicted all entities correctly
}}

\vspace{0.5em}

\fbox{\parbox{0.95\textwidth}{
\textbf{Example 4: Historical Sovereign State} (country vs organization)\\[0.5em]
\textit{``The [attack] was part of the strategic bombing campaign waged by the United States of America against military and civilian targets and population centers of the \underline{Empire of Japan} during the Japan home islands campaign in the closing stages of World War II.''}\\[0.5em]
\textbf{Ground Truth:} Empire of Japan $\rightarrow$ COUNTRY\\[0.3em]
\textbf{Model Errors:}\\
\quad GLiNER: Labeled as ORGANIZATION instead of COUNTRY\\
\quad UniNER: Mixed predictions (includes ORGANIZATION in addition to COUNTRY)\\
\quad JPT: Predicted all entities correctly
}}
\caption{Examples from CrossNER Politics where our model correctly identifies fine-grained entity types. GLiNER confuses adjacent categories (POLITICAL PARTY/ORGANIZATION, POLITICIAN/PERSON, COUNTRY/ORGANIZATION). Because UniNER is inferred per entity type, it is more prone to over-prediction (multiple types per mention). Our model leverages type definitions to resolve these distinctions.}
\label{fig:disambiguation-examples}
\end{figure*}

\begin{table*}[t]
\centering
\small
\resizebox{\textwidth}{!}{%
\begin{tabular}{llp{6cm}lll}
\toprule
\textbf{Error Type} & \textbf{Dataset} & \textbf{Text (excerpt)} & \textbf{Gold} & \textbf{Pred.} & \textbf{Analysis} \\
\midrule
Type Confusion & CrossNER-AI & ``NIST also differs from \underline{Bilingual evaluation understudy} in its calculation\ldots'' & \textsc{metrics} & \textsc{conference} & Acronym ambiguity \\
Type Confusion & CrossNER-Politics & ``The provinces ceded to \underline{Augustus} for that ten-year period\ldots'' & \textsc{person} & \textsc{politician} & Role vs.\ entity \\
Type Confusion & CrossNER-Music & ``\ldots on \underline{The Clash}'s \underline{London Calling}\ldots'' & \textsc{album} & \textsc{location} & Title--location ambiguity \\
\midrule
Missed Entity & CrossNER-AI & ``Rethink Robotics introduced Baxter as an \underline{industrial robot}\ldots'' & \textsc{product} & (none) & Descriptive form \\
Missed Entity & CrossNER-Politics & ``growth in the \underline{Melanesian} countries of Solomon Islands\ldots'' & \textsc{location} & (none) & Adjectival reference \\
\midrule
Over-prediction & CrossNER-AI & ``A \underline{frame language} is a technology used for knowledge representation\ldots'' & (none) & \textsc{prog.\ lang.} & Over-generalization \\
Over-prediction & CrossNER-Politics & ``reforms to the \underline{ALP} under Gough Whitlam\ldots'' & (none) & \textsc{pol.\ party} & Unannotated entity \\
\bottomrule
\end{tabular}%
}
\caption{Representative \textbf{JPT-4B} error instances with context excerpts, gold labels, predictions, and diagnostics.}
\label{tab:failure-examples}
\end{table*}

\end{document}